\documentclass[12pt]{article}

\usepackage[utf8]{inputenc}
\usepackage{amsmath}
\usepackage{amssymb}
\usepackage{amsthm}
\usepackage{xcolor}
\usepackage{graphicx}
\graphicspath{{N:/MatMoDatPrivate/martin09/linux/Projects/Uncertainty_Deep_Learning/Pics/}}
\usepackage[font=scriptsize,labelfont=bf]{caption}
\makeatletter
\def\input@path{{N:/MatMoDatPrivate/martin09/linux/Projects/Uncertainty_Deep_Learning/Pics/}}
\makeatother

\usepackage{subcaption}
\usepackage{pgfplots}

\newcommand{\Fi}{\mathbb{F}}
\newcommand{\EE}{\mathbb{E}}
\newcommand{\KL}{\mathrm{KL}}
\newcommand{\tr}{\mathrm{tr}}
\newcommand{\eps}{\varepsilon}

\newcommand{\argmax}{\mathrm{arg}\,\max}

\title{Inspecting adversarial examples using the Fisher information}

\author{Jörg Martin, Clemens Elster \\
Physikalisch-Technische Bundesanstalt (PTB)}

\begin{document}
\maketitle

\begin{abstract}
Adversarial examples are slight perturbations that are designed to fool artificial neural networks when fed as an input. In this work the usability of the Fisher information for the detection of such adversarial attacks is studied. We discuss various quantities whose computation scales well with the network size, study their behavior on adversarial examples and show how they can highlight the importance of single input neurons, thereby providing a visual tool for further analyzing (un-)reasonable behavior of a neural network. The potential of our methods is demonstrated by applications to the MNIST, CIFAR10 and Fruits-360 datasets. 
\end{abstract}
\section{Introduction}

It is occasionally said that the term \emph{artificial neural network} (ANN) is poorly chosen as the underlying mechanism has hardly anything in common with biological neurons \cite{Bengio2015}. The existence of adversarial examples seems to back this criticism: A tiny change to the input to an ANN, often on an almost  imperceptible scale for a human, can lead to a fundamentally different output provided the dimensionality of the input space is high enough \cite{Goodfellow2014}. It seems as if neural networks are only working as we expect them to do on a small subdomain of the actual input space, while deviations might lead to an unforeseen outcome. This phenomenon is not only discomforting from a conceptual perspective but can in fact become safety-relevant \cite{Amodei2016}, especially since even physical, three-dimensional objects can be equipped with an adversarial property \cite{Kurakin2016, Athalye2017}. Techniques that can raise a red flag on suspect behavior have therefore become a relevant topic in current research about adversarial examples \cite{Akhtar2018}. We here show how Fisher's information matrix can be transformed into a tool for such purposes. Several quantities are proposed that measure the size of the latter and allow for adversarial detection while preserving scalability. Using a simple stochastic model for the input of the network these quantities can be used to highlight input nodes of specific importance which can serve as a further insight to a possible ``unreasonable'' behavior of an ANN.  

Since their discovery by Szegedy et al. \cite{Szegedy2013} a variety of techniques for possible adversarial attacks have been identified, such as the Fast gradient sign method (FGSM) \cite{Goodfellow2014}, methods involving momentum \cite{Dong2018}, C\&W attacks \cite{Carlini2017Towards} or DeepFool \cite{Moosavi2016} to name only a few. These methods slightly disturb the input into a certain direction that is crafted to mislead the neural network  - this can be achieved even without access to the weights \cite{Chen2017}. In the case of input images it might be enough to disturb only one pixel \cite{Su2019}. Methods designed for the defense against such attacks either consider robustification of the ANN \cite{Papernot2016, Madry2017, Zantedeschi2017, Shaham2018} or present, such as this article, methods to detect adversarial inputs. In the latter case inputs that would count as adversarial under these criteria can then be re-used in training to robustify and regularize the neural network \cite{Goodfellow2014}. The unmasking of adversarial examples is often achieved via addition of a certain substructure to the network \cite{Metzen2017, Grosse2017} or an additional classifier \cite{Lu2017, Xu2017}. Other methods, such as the one presented in this article, use statistical methods \cite{Li2017, Liang2017, Feinman2017, Smith2018}. While it seems, at least in certain situations, that there is an upper bound on the effectiveness of adversarial attacks \cite{Raghunathan2018}, the defense against them remains so far an unresolved issue in the application of neural networks \cite{Carlini2017}. This article complements the literature in two ways: 
\begin{itemize}
\item We study adversarial detection based on the \emph{Fisher information matrix}, which measures the average magnitude of the derivative of the log-likelihood evaluated at the learned parameter, namely the weights and biases of the network. 
\item  In a second step we derive from the proposed quantities a vector that indicates for each input node its importance measured in terms of the Fisher information when exposed to hypothetical noise. This allows, say for applications such as image classification, a \emph{highlighting of relevant areas} that can be used to further examine the (un-)reasonableness of a network output.
\end{itemize}
In much of the literature adversarial examples are postulated as lying outside the ``manifold of natural data'' \cite{Gardner2015, Feinman2017, Smith2018}. The Fisher information puts us in a kind of a dual view: We consider the curvature of the \emph{statistical manifold} (in a loose sense, \emph{not} in the one of differential geometry) that arises for an input at the previously learned parameter values, thus judging how well the input fits with the learned network structure. In \cite{Shi2019, Zhao2019} the authors considered somewhat related ideas to the ones persued here, but by treating the input of the network as parameter. We here treat the actual parameters of the network (usually weights and biases) as the ones considered by the Fisher information. In \cite{Nayebi2017} the authors consider the same Fisher information as we do and use a similar objects as in \eqref{eq:FisherForm} below to propose a robustified learning approach. 

The structure of this article is as follows: In Section \ref{sec:detection} we recall the definition and meaning of the Fisher information and then propose several quantities whose computational complexity scales linearly with the network size. The effect of an adversarial attack on the MNIST dataset is discussed for these quantities (Figure \ref{fig:MNIST_Hist_ROC}). In Section \ref{sec:visualization} we show how the addition of a hypothetical noise to the input allows for derivation of a quantity that can visualize the influence of single input nodes on the Fisher information.  Section \ref{sec:experiments} describes the application of the presented methods from Section \ref{sec:detection} and \ref{sec:visualization} to MNIST, CIFAR10 and Fruits-360.

\section{Detecting adversarial attacks with the Fisher information}
\label{sec:detection}
\subsection{A short recap on the Fisher information}
Consider data $\mathcal{D}$ that follows some sampling distribution $p(\mathcal{D} |\theta)$ parametrized
by a $p$-dimensional parameter vector $\theta$. The Fisher information for a given value of $\theta$ equals \cite{Bishop2006}
\begin{align}
\label{eq:FisherInformation}
\begin{aligned}
\Fi_\theta &= \EE_{\mathcal{D}\sim p(\mathcal{D}|\theta)}\left[\nabla_\theta \log p(\mathcal{D}|\theta) \nabla_\theta^T \log p(\mathcal{D}|\theta)\right] \\
&=- \EE_{\mathcal{D}\sim p(\mathcal{D}|\theta)}\left[\nabla_\theta \nabla_\theta^T \log p(\mathcal{D}|\theta)\right] \,.
\end{aligned}
\end{align}
(provided the  log-likelihood is sufficiently smooth, which we assume for simplicity throughout this article). This matrix measures the averaged local curvature of the log-likelihood
$\log p(\mathcal{D}|\theta)$. Occasionaly one thinks of the parametrized distributions $p(\mathcal{D}|\theta)$
as lying on a (statistical) manifold with a distance notion given by the
Kullback-Leibler-divergence \cite{Amari2012}. If we consider a small pertubation of $\theta$ in direction of a $p$ dimensional vector $v$, say $\theta + \eps\cdot v$ then one checks via Taylor
expansion that the Kullback-Leibler divergence behaves as
\begin{align}
\label{eq:KLTaylor}
\KL(p(\mathcal{D}|\theta)\,|\!\|\, p(\mathcal{D}|\theta+\eps\cdot v)) = \frac{1}{2}\eps^2 v^T \Fi_\theta v + \mathcal{O}(\eps^3)
\end{align}
Thus, $\Fi_\theta$ can be seen as the metric that quantifies how much information
can locally be gained about the parameter $\theta$ (the stat. manifold is in fact equipped with $\Fi_\theta$ as metric \cite{Amari2012}). A high value of $\Fi_\theta$ indicates that the
parameter is not well-adjusted to the data or, from a different point of view,
the ability to draw information about $\theta$ from the data is pretty high. The first perspective motivates to see $\Fi_\theta$ as a measure about the ``unusualness'' of the data if we assume that our parameter vector $\theta$ is correct. If we can thus somehow measure the size of $\Fi_\theta$ we can use this to detect unusual data such as adversarial examples. 

Alas, the dimension of $\Fi_\theta$ is $p\times p$ where the parameter dimension $p$ is even for a rather small network already of the order $10^6$. Computing the full matrix $\Fi_\theta$ thus soon gets intractable. In the next section we will discuss quantities that measure the size of $\Fi_\theta$ without requiring us to compute the full matrix \eqref{eq:FisherInformation}. 

\subsection{Quantifying the Fisher information}
\label{subsec:scalable_quantities}

\begin{figure}
\begin{subfigure}[t]{0.5\textwidth}
\includegraphics[scale=0.5]{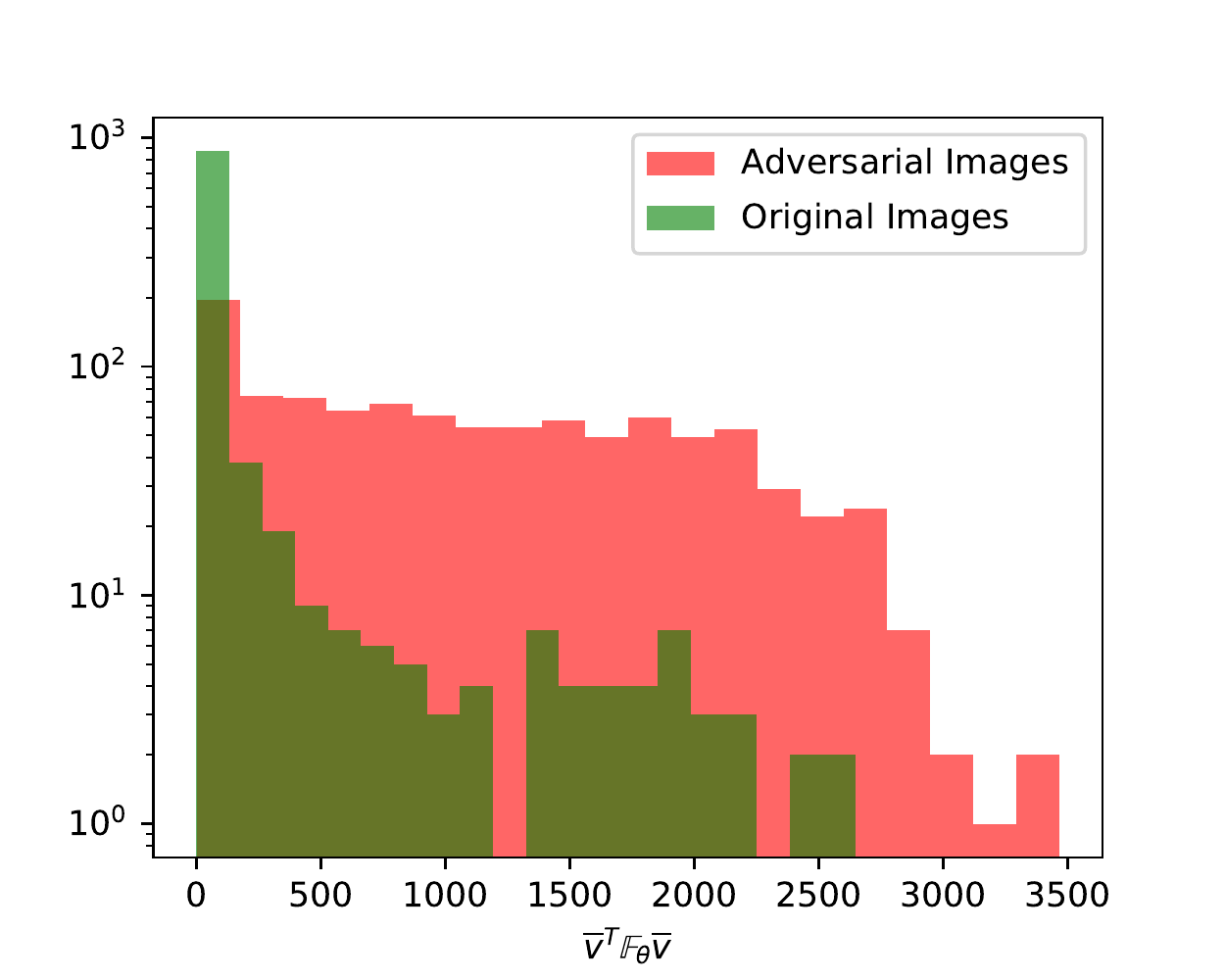}
\end{subfigure}
\begin{subfigure}[t]{0.5\textwidth}
\begin{tikzpicture}

\begin{axis}[
font=\fontsize{6}{6},
height=2 in,
legend cell align={left},
legend style={at={(0.97,0.03)}, anchor=south east, draw=none, fill=none},
tick align=outside,
tick pos=left,
width=2.4 in,
x grid style={white!69.01960784313725!black},
xlabel={\(\displaystyle \mathrm{FPR}\)},
xmin=-0.05, xmax=1.05,
xtick style={color=black},
y grid style={white!69.01960784313725!black},
ylabel={\(\displaystyle \mathrm{TPR}\)},
ymin=-0.05, ymax=1.05,
ytick style={color=black}
]
\addplot [thick, black]
table {%
0 0
0 0.00515463917525773
0 0.0360824742268041
0.00485436893203883 0.0360824742268041
0.00485436893203883 0.185567010309278
0.00970873786407767 0.185567010309278
0.00970873786407767 0.391752577319588
0.0145631067961165 0.391752577319588
0.0145631067961165 0.561855670103093
0.0194174757281553 0.561855670103093
0.0194174757281553 0.603092783505155
0.029126213592233 0.603092783505155
0.029126213592233 0.61340206185567
0.0436893203883495 0.61340206185567
0.0436893203883495 0.639175257731959
0.0485436893203883 0.639175257731959
0.0485436893203883 0.65979381443299
0.0533980582524272 0.65979381443299
0.0533980582524272 0.664948453608247
0.058252427184466 0.664948453608247
0.058252427184466 0.814432989690722
0.0679611650485437 0.814432989690722
0.0679611650485437 0.819587628865979
0.0825242718446602 0.819587628865979
0.0825242718446602 0.829896907216495
0.087378640776699 0.829896907216495
0.087378640776699 0.860824742268041
0.0970873786407767 0.860824742268041
0.0970873786407767 0.871134020618557
0.116504854368932 0.871134020618557
0.116504854368932 0.891752577319588
0.131067961165049 0.891752577319588
0.131067961165049 0.896907216494845
0.145631067961165 0.896907216494845
0.145631067961165 0.907216494845361
0.160194174757282 0.907216494845361
0.160194174757282 0.912371134020619
0.174757281553398 0.912371134020619
0.174757281553398 0.917525773195876
0.194174757281553 0.917525773195876
0.194174757281553 0.922680412371134
0.199029126213592 0.922680412371134
0.199029126213592 0.927835051546392
0.20873786407767 0.927835051546392
0.20873786407767 0.93298969072165
0.213592233009709 0.93298969072165
0.213592233009709 0.938144329896907
0.223300970873786 0.938144329896907
0.223300970873786 0.943298969072165
0.233009708737864 0.943298969072165
0.233009708737864 0.948453608247423
0.247572815533981 0.948453608247423
0.247572815533981 0.95360824742268
0.262135922330097 0.95360824742268
0.262135922330097 0.963917525773196
0.281553398058252 0.963917525773196
0.281553398058252 0.969072164948454
0.364077669902913 0.969072164948454
0.364077669902913 0.974226804123711
0.368932038834951 0.974226804123711
0.368932038834951 0.984536082474227
0.422330097087379 0.984536082474227
0.422330097087379 0.989690721649485
0.441747572815534 0.989690721649485
0.441747572815534 0.994845360824742
0.495145631067961 0.994845360824742
0.495145631067961 1
0.990291262135922 1
1 1
};
\addlegendentry{$\mathrm{tr}\, \mathbb{F}_\theta$}
\addplot [thick, blue, dashed]
table {%
0 0
0 0.00531914893617021
0 0.117021276595745
0.00471698113207547 0.117021276595745
0.00471698113207547 0.132978723404255
0.00943396226415094 0.132978723404255
0.00943396226415094 0.164893617021277
0.0141509433962264 0.164893617021277
0.0141509433962264 0.186170212765957
0.0188679245283019 0.186170212765957
0.0188679245283019 0.271276595744681
0.0235849056603774 0.271276595744681
0.0235849056603774 0.313829787234043
0.0283018867924528 0.313829787234043
0.0283018867924528 0.377659574468085
0.0330188679245283 0.377659574468085
0.0330188679245283 0.441489361702128
0.0471698113207547 0.441489361702128
0.0471698113207547 0.473404255319149
0.0518867924528302 0.473404255319149
0.0518867924528302 0.50531914893617
0.0566037735849057 0.50531914893617
0.0566037735849057 0.531914893617021
0.0613207547169811 0.531914893617021
0.0613207547169811 0.553191489361702
0.0660377358490566 0.553191489361702
0.0660377358490566 0.595744680851064
0.0707547169811321 0.595744680851064
0.0707547169811321 0.664893617021277
0.0754716981132075 0.664893617021277
0.0754716981132075 0.702127659574468
0.0849056603773585 0.702127659574468
0.0849056603773585 0.728723404255319
0.089622641509434 0.728723404255319
0.089622641509434 0.776595744680851
0.0943396226415094 0.776595744680851
0.0943396226415094 0.781914893617021
0.0990566037735849 0.781914893617021
0.0990566037735849 0.792553191489362
0.10377358490566 0.792553191489362
0.10377358490566 0.824468085106383
0.108490566037736 0.824468085106383
0.108490566037736 0.840425531914894
0.117924528301887 0.840425531914894
0.117924528301887 0.861702127659574
0.127358490566038 0.861702127659574
0.127358490566038 0.877659574468085
0.155660377358491 0.877659574468085
0.155660377358491 0.893617021276596
0.160377358490566 0.893617021276596
0.160377358490566 0.920212765957447
0.193396226415094 0.920212765957447
0.193396226415094 0.925531914893617
0.19811320754717 0.925531914893617
0.19811320754717 0.930851063829787
0.212264150943396 0.930851063829787
0.212264150943396 0.936170212765957
0.216981132075472 0.936170212765957
0.216981132075472 0.941489361702128
0.240566037735849 0.941489361702128
0.240566037735849 0.946808510638298
0.245283018867925 0.946808510638298
0.245283018867925 0.957446808510638
0.264150943396226 0.957446808510638
0.264150943396226 0.968085106382979
0.268867924528302 0.968085106382979
0.268867924528302 0.973404255319149
0.292452830188679 0.973404255319149
0.292452830188679 0.978723404255319
0.297169811320755 0.978723404255319
0.297169811320755 0.984042553191489
0.391509433962264 0.984042553191489
0.391509433962264 0.99468085106383
0.443396226415094 0.99468085106383
0.443396226415094 1
0.990566037735849 1
1 1
};
\addlegendentry{$v^T\mathbb{F}_\theta v$}
\addplot [thick, white!50.19607843137255!black, dashed]
table {%
0 0
0 0.00458715596330275
0 0.188073394495413
0.00549450549450549 0.188073394495413
0.00549450549450549 0.201834862385321
0.010989010989011 0.201834862385321
0.010989010989011 0.298165137614679
0.0164835164835165 0.298165137614679
0.0164835164835165 0.522935779816514
0.021978021978022 0.522935779816514
0.021978021978022 0.564220183486238
0.0274725274725275 0.564220183486238
0.0274725274725275 0.568807339449541
0.032967032967033 0.568807339449541
0.032967032967033 0.573394495412844
0.0384615384615385 0.573394495412844
0.0384615384615385 0.697247706422018
0.043956043956044 0.697247706422018
0.043956043956044 0.711009174311927
0.0494505494505494 0.711009174311927
0.0494505494505494 0.729357798165138
0.0549450549450549 0.729357798165138
0.0549450549450549 0.73394495412844
0.0659340659340659 0.73394495412844
0.0659340659340659 0.756880733944954
0.0714285714285714 0.756880733944954
0.0714285714285714 0.779816513761468
0.0769230769230769 0.779816513761468
0.0769230769230769 0.798165137614679
0.0824175824175824 0.798165137614679
0.0824175824175824 0.807339449541284
0.0879120879120879 0.807339449541284
0.0879120879120879 0.811926605504587
0.0934065934065934 0.811926605504587
0.0934065934065934 0.81651376146789
0.10989010989011 0.81651376146789
0.10989010989011 0.825688073394495
0.131868131868132 0.825688073394495
0.131868131868132 0.839449541284404
0.137362637362637 0.839449541284404
0.137362637362637 0.844036697247706
0.153846153846154 0.844036697247706
0.153846153846154 0.862385321100917
0.159340659340659 0.862385321100917
0.159340659340659 0.880733944954128
0.164835164835165 0.880733944954128
0.164835164835165 0.885321100917431
0.175824175824176 0.885321100917431
0.175824175824176 0.889908256880734
0.181318681318681 0.889908256880734
0.181318681318681 0.894495412844037
0.186813186813187 0.894495412844037
0.186813186813187 0.912844036697248
0.197802197802198 0.912844036697248
0.197802197802198 0.917431192660551
0.203296703296703 0.917431192660551
0.203296703296703 0.926605504587156
0.21978021978022 0.926605504587156
0.21978021978022 0.931192660550459
0.230769230769231 0.931192660550459
0.230769230769231 0.935779816513762
0.269230769230769 0.935779816513762
0.269230769230769 0.944954128440367
0.28021978021978 0.944954128440367
0.28021978021978 0.954128440366973
0.285714285714286 0.954128440366973
0.285714285714286 0.958715596330275
0.296703296703297 0.958715596330275
0.296703296703297 0.963302752293578
0.302197802197802 0.963302752293578
0.302197802197802 0.972477064220184
0.340659340659341 0.972477064220184
0.340659340659341 0.977064220183486
0.434065934065934 0.977064220183486
0.434065934065934 0.981651376146789
0.461538461538462 0.981651376146789
0.461538461538462 0.990825688073395
0.675824175824176 0.990825688073395
0.675824175824176 0.995412844036697
0.884615384615385 0.995412844036697
0.884615384615385 1
1 1
};
\addlegendentry{$\overline{v}^T\mathbb{F}_\theta \overline{v}$}
\addplot [thick, green!50.19607843137255!black, dotted]
table {%
0 0
1 1
};
\addlegendentry{$\mathrm{Diagonal}$}
\end{axis}

\end{tikzpicture}
\end{subfigure}
\caption{\textit{Left}: Distribution of the quantity $\overline{v}^T \mathbb{F}_\theta \overline{v}$ from \eqref{eq:NormalizedFisherForm} for 1000 MNIST images (green) and their modifications trough an adversarial attack (red). For details on the adversarial attack and the computation of $\overline{v}^T \mathbb{F}_\theta \overline{v}$ compare Section \ref{subsec:MNIST}. \textit{Right}: 
ROC for adversarial detection based on the three quantities derived from the Fisher information in Section \ref{sec:detection} for the MNIST dataset. The AUC (area under the ROC curve) is given by 0.94 for the trace of Fisher information  - \eqref{eq:FisherTrace}, 0.92 for the quadratic form in \eqref{eq:FisherForm} and 0.93 for the normalized version from \eqref{eq:NormalizedFisherForm}. }
\label{fig:MNIST_Hist_ROC}
\end{figure}
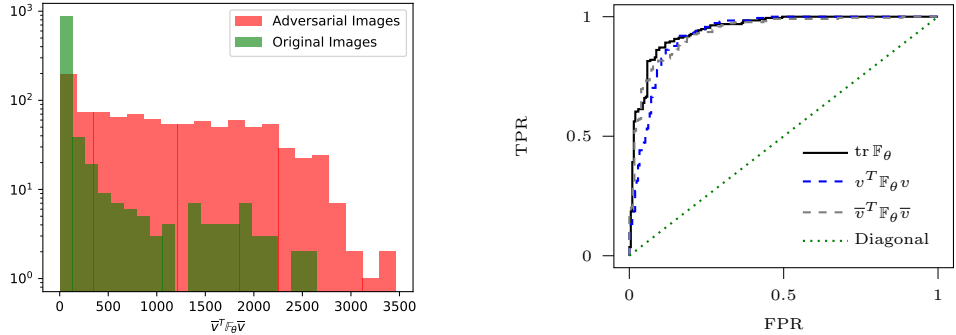

In machine learning, one often tries to learn a parametrized mapping
$f_\theta$  from the data. In the case of a neural network, which is the only scenario considered in this article, $\theta$ will contain all weights and biases for all connections of the network and $f_\theta$ will map an input, say an image, to an output, such as a vector obtained by an application of a softmax. To learn $\theta$ assume we are given a collection of $N$ observations of inputs $X$ and corresponding targets $Y$. In the notation of the previous subsection $(X, Y )$ can be considered as $N$ realizations $(x,y)$ of the random variable $\mathcal{D}$. The aim is now to find $\theta$ such that $f_\theta (x) \approx y$ for each pair of the $N$ observations $(x,\, y)$ from $(X,\, Y)$. To embed this task into a mathematical framework one imposes a probability model which is somehow linked with the mapping $f_\theta$. A common choice would be to consider $f_\theta(x)$ as the mean of a Gaussian distribution from which $y$ is drawn (conditional on $x$). For classification problems in $C$ classes, where the target $y$ will lie in one of these classes, one usually chooses the output $f_\theta(x)$ to be a $C$-dimensional vector that
sums to 1, as a kind of generalized logistic regression \cite{Bishop2006}. We will here solely focus on this second case as it seems to be the one that is mostly treated concerning adversarial examples.\footnote{Similar formulas to the ones derived in Section \ref{subsec:scalable_quantities} and \ref{sec:visualization} can however also be derived for Gaussian likelihoods.}

Suppose therefore that for an input $x$ the components of the $C$-dimensional vector $f_\theta (x) = (f_\theta^c(x))_{c=1,...,C}$ are non-negative, sum to
1 and are interpreted as the probabilities for class membership. If we treat
$x$ as deterministic, that is we condition on $x$, we can write the Fisher information in \eqref{eq:FisherInformation}  as
\begin{align}
\label{eq:FisherInformationCategorical}
\Fi_\theta=\sum_{c=1}^C f_\theta^c(x) \cdot \nabla_\theta \log f_\theta^c(x) \nabla_\theta^T \log f_\theta^c(x)
 = \sum_{c=1}^C \nabla_\theta f_\theta^c(x) \nabla_c^T\log f_\theta(x)\,.
\end{align}
As already pointed out above, even for relatively small neural networks a full computation of $\Fi_\theta$ soon becomes intractable. However, if we are only interest in some quantity that measures the size of \eqref{eq:FisherInformationCategorical} we can use the trace
\begin{align}
\label{eq:FisherTrace}
\tr \Fi_\theta = \sum_{i=1}^p \sum_{c=1}^C \partial_{\theta_i} f_\theta^c(x) \partial_{\theta_i} \log f_\theta^c(x)
\end{align}
which can be computed in $O(C\cdot p)$ by
using backpropagation. Using the class reduction we introduce in \eqref{eq:2class} below this can be reduced to $O(p)$ so that the computational complexity scales linearly with the network size. 

Note that we can write the trace of $\Fi_\theta$ as $\sum_{i=1}^p e_i^T \Fi_\theta e_i$ which can be read as an average over the changes in the Kullback-Leibler distance w.r.t to each parameter, compare the expansion \eqref{eq:KLTaylor}. This motivates an alternative
quantity to \eqref{eq:FisherTrace}. Instead of averaging over a full orthonormal basis we might pick one
specific direction and measure, using \eqref{eq:FisherInformationCategorical}, the quadratic form
\begin{align}
\label{eq:FisherForm}
v^T \Fi_\theta v = \sum_{c=1}^C v^T \nabla_\theta f_\theta^c(x) \cdot  v^T \nabla_\theta \log f_\theta^c(x)
\end{align}
We here take a direction vector that is dependent on the parameter $\theta$ and the considered datapoint $(x, y)$, namely
\begin{align}
\label{eq:DirectionNLLLoss}
v=\lambda \cdot \nabla_\theta \log p(y|x,\theta) 
\end{align}
If we choose the scaling $\lambda>0$ to be the learning rate than according to \eqref{eq:KLTaylor}
the object \eqref{eq:FisherForm} approximates the Kullback-Leibler divergence corresponding
to a single learning step in a stochastic gradient descent training (up to a factor 1/2).\footnote{We should point out that taking $v$ equal to the natural gradient instead would transform \eqref{eq:FisherForm} into the Fisher kernel \cite{Jaakkola1999}, whose computation is however practically unfeasible for most neural networks.}  We will see in Section \ref{sec:visualization} below that \eqref{eq:FisherForm} allows for the effective computation of a visualization
tool that allows to highlight input components that are of particular importance. In practice $y$, as used in \eqref{eq:DirectionNLLLoss}, is often unknown, so that we here pick the predicted
class label, that is $\hat{y} = \argmax_{c=1,...,C} f_\theta^c (x)$ . This prediction can also be used to give an approximation to summations over the set of classes such as in \eqref{eq:FisherTrace} or in \eqref{eq:FisherForm} by replacing the $C$-dimensional vector $(f_\theta^c (x))_{c=1,...,C}$ by a two-dimensional
vector
\begin{align}
\label{eq:2class}
(f_\theta^{\hat{y}} (x), \sum_{c\neq \hat{y}} f_\theta^c(x)) = (f_\theta^{\hat{y}}(x), 1 - f_\theta^{\hat{y}} (x))\,.
\end{align}
This corresponds to a classification with two labels, namely $\hat{y}$ or \emph{not} $\hat{y}$ . Instead of taking \eqref{eq:FisherForm} one might also pick the normalized object
\begin{align}
\label{eq:NormalizedFisherForm}
\overline{v}^T \Fi_\theta \overline{v}
\end{align}
where $\overline{v} = v/\|v\|$ , that ignores the magnitude of $v$. This object can be seen as solely measuring the curvature of the statistical manifold at $\theta$. We should point out once more that whenever we speak of curvature this should be understood in loose sense and \emph{not} in the one of differential geometry.  Computing the actual, say Ricci-, curvature for the statistical manifold of a neural network is probably unfeasible. We will see below that \eqref{eq:NormalizedFisherForm} and \eqref{eq:FisherForm} perform equally in the detection of adversarial examples. Using backpropagation and an eventual class reduction as in \eqref{eq:2class} all quantities we introduced here, namely \eqref{eq:FisherTrace}, \eqref{eq:FisherForm} and \eqref{eq:NormalizedFisherForm} are of a computational complexity that scales linear with the network size. For the two quadratic forms \eqref{eq:FisherForm} and \eqref{eq:NormalizedFisherForm} there is even an approximation that spares us the backpropagation (for a given $v$). In fact, note that in \eqref{eq:FisherForm} the derivative w.r.t $\theta$ only appears as a directional derivative, which allows us to approximate
\begin{align}
\label{eq:FiniteDifference}
v^T \nabla_\theta f^c_\theta(x) \simeq (f_{\theta+\eps' \cdot v}-f_{\theta})/\eps' \,,
\end{align}
and similar for $\log f_\theta(x)$, where $\eps' > 0$ is small. We will refer to this as the
\emph{finite difference approximation} - this will turn out quite handy when we introduce the
visualization technique in the following section.

The left-hand side of Figure \ref{fig:MNIST_Hist_ROC} shows the evaluation of \eqref{eq:NormalizedFisherForm} for 1000 randomly chosen MNIST images, before (green) and after (red) an adversarial attack. For the details on the adversarial attack we refer to Section \ref{sec:experiments} below. We used the numerical approximation from \eqref{eq:FiniteDifference}. Note, that the distribution of $\overline{v}^T\Fi_\theta \overline{v}$ is spread out to larger values as the statistical manifold will be more curvy for images that do not suit the learned parameters such as adversarial examples. Agreeing on a certain critical value we can thus classify an image as adversarial once $\overline{v}^T\Fi_\theta \overline{v}$ exceeds this threshold. Varying the later yields the receiver operating characteristic (ROC) that plots the true positive rate (TPR) versus the false positive rate (FPR). In the right  - hand side of Figure \ref{fig:MNIST_Hist_ROC} this is done for the MNIST dataset and all quantities introduced in this section, namely the trace \eqref{eq:FisherTrace}, the quadratic form \eqref{eq:FisherForm} and its normalized counterpart \eqref{eq:NormalizedFisherForm}. All three curves have an area under of curve (AUC) of around 0.92 (0.5 would be guessing, 1.0 perfect detection). For more details, such as the used adversarial attack, we refer to Section \ref{sec:experiments} below.

\section{Fisher information sensitivity}
\label{sec:visualization}
To enhance the information we can draw from the quantities \eqref{eq:FisherForm} and \eqref{eq:NormalizedFisherForm} let us make the input to the neural network stochastic by introducing a univariate noise variable $\xi\sim \mathcal{N}(0,1)$, independent of the hereto used framework, and defining
\begin{align}
\label{eq:xeps}
x^{\eps,\eta} = x+\eps \, \xi\cdot \eta
\end{align}
where $\eps>0$ and where $\eta$ has the same dimensionality as $x$. In a way this is similar to model used in \cite{Liang2017}. The Fisher information for this noisified input, once more conditional on $x$, can then  be expressed as
\begin{align*}
\Fi^{\eps,\eta}_\theta &=  \sum_{c=1}^C \EE_{x^{\eps,\mu}}[\nabla_\theta f_\theta^c(x^{\eps,\mu}) \nabla_c^T\log f_\theta(x^{\eps,\mu})] \\
&=  \Fi_\theta  + 0 +\frac{1}{2} \eps^2 \sum_{c=1}^C \sum_{i,j=1}^N \eta_i \nabla_\theta \partial_{x_i} f_\theta^c(x) \nabla_c^T \partial_{x_j}\log f_\theta(x) \eta_j + \mathcal{O}(\eps^3) \,,
\end{align*}
where we used the zero mean of the components of $\xi$ to drop the first order term. The second order term is practically incomputable for neural networks of any interesting size as it involves second order derivatives. However, if measure the quadratic form as in \eqref{eq:FisherForm}, or a normalized one as in \eqref{eq:NormalizedFisherForm}, we can write
\begin{align}
\label{eq:ExpansionFisherDifference}
v^\top \Fi^{\eps,\eta}_\theta v = v^\top \Fi_\theta v  + \eta^\top \delta_v \Fi_\theta \eta
\end{align}
where 
\begin{align}
\label{eq:FullFisherDifference}
\delta_v \Fi_\theta = \sum_{c=1}^C \nabla_x (v^\top \nabla_\theta f_\theta^c(x)) \cdot \nabla_x^\top(v^\top \nabla_\theta \log f_\theta^c(x))
\end{align}
Making use of the finite difference approximation from \eqref{eq:FiniteDifference} we can replace the inner gradients by finite differences in the direction of $v$, so that we only have to compute a first order derivative which can be done efficiently via backpropagation. From  \eqref{eq:FullFisherDifference} one might compute various objects depending on the choice of $\eta$. We here take a straightforward approach, that scales well with large input dimensions. Namely, for each $i=1,\ldots,N$ take $\eta = e_i$ and compute in \eqref{eq:ExpansionFisherDifference} the second term as
\begin{align}
\label{eq:FIS}
e_i^T \delta_v \Fi_\theta e_i= \sum_{c=1}^C \partial_{x_i}(v^\top \nabla_\theta f_\theta^c(x)) \cdot \partial_{x_i}(v^\top \nabla_\theta \log f_\theta^c(x))
\end{align}
Doing this for every $i$ we obtain a vector $(e_i^T \delta_v \Fi_\theta e_i)_{i=1,\ldots,N}$, which we call the \emph{Fisher information sensitivity} (FIS). We can read the FIS as a measure of the importance of the $i$th input node. Via the finite difference approximation from \eqref{eq:FiniteDifference}, backpropagation and an eventual class number reduction as in \eqref{eq:2class} we can efficiently compute this vector in \emph{one} backward pass. Let us point out that normalizing $v$ as in \eqref{eq:NormalizedFisherForm} for the FIS only plays a role if we want to consider the FIS on an absolute scale. Once we only care about the relative proportions between different $i$ a possible normalization factor of $v$ looses any relevance. 

\section{Experiments}
\label{sec:experiments}

In this section we test the detection of adversarial examples on the MNIST, CIFAR10 and Fruits-360 dataset using the quantities \eqref{eq:FisherTrace}, \eqref{eq:FisherForm} and \eqref{eq:NormalizedFisherForm} from Section \ref{sec:detection}. All images were scaled to the range $[0,1]$. The adversarial attacks in this section were produced using the MI-FGSM\cite{Dong2018} with $T=10,u=1.0$ and varying parameter $\epsilon$. An application of the FGSM \cite{Goodfellow2014} yielded comparable results. 

The results in this section are presented by the receiver operating characteristic which was computed by comparing the values of \eqref{eq:FisherTrace}, \eqref{eq:FisherForm} and \eqref{eq:NormalizedFisherForm} on adversarial and original images. Using the FIS from Section \ref{sec:visualization} we visualize the importance of each input pixel for a few test samples. 
\subsection{MNIST}
\label{subsec:MNIST}

\begin{figure}
\begin{subfigure}[c]{0.5\textwidth}
\includegraphics[scale=0.4]{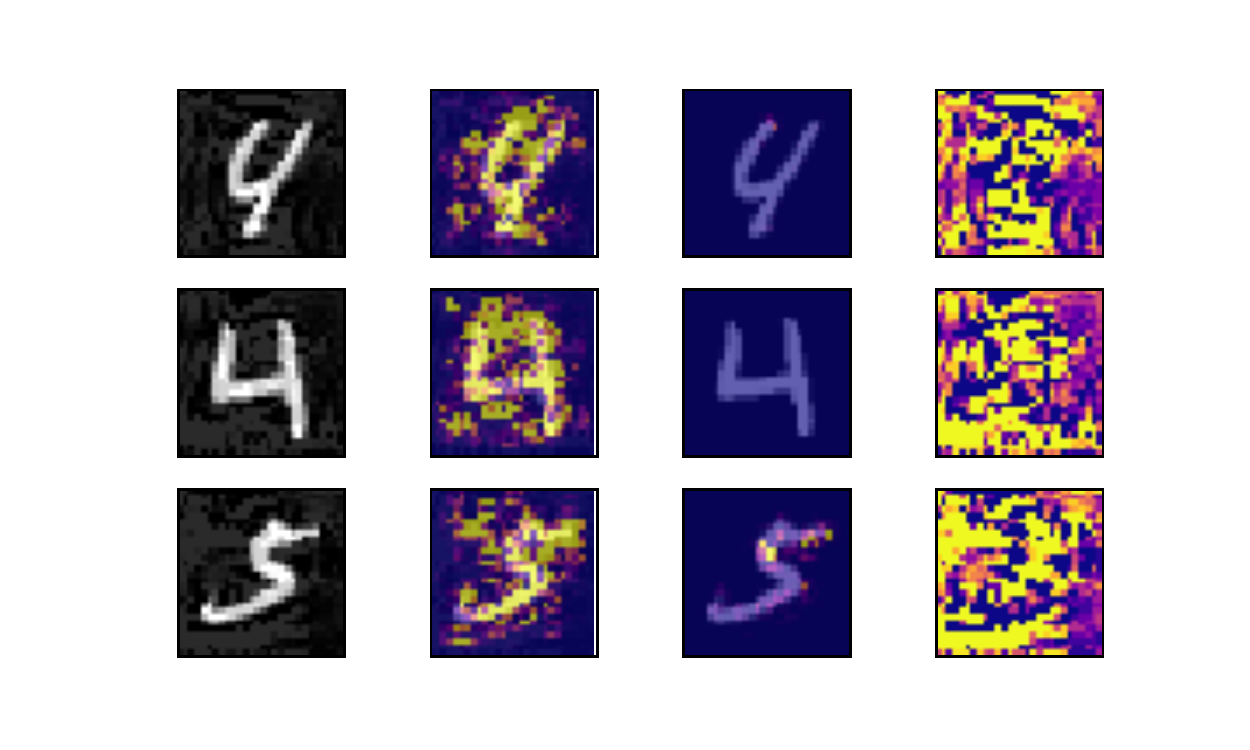}
\subcaption{FIS: Same color scales.}
\label{fig:MNIST_Diff}
\end{subfigure}
\begin{subfigure}[c]{0.5\textwidth}
\includegraphics[scale=0.4]{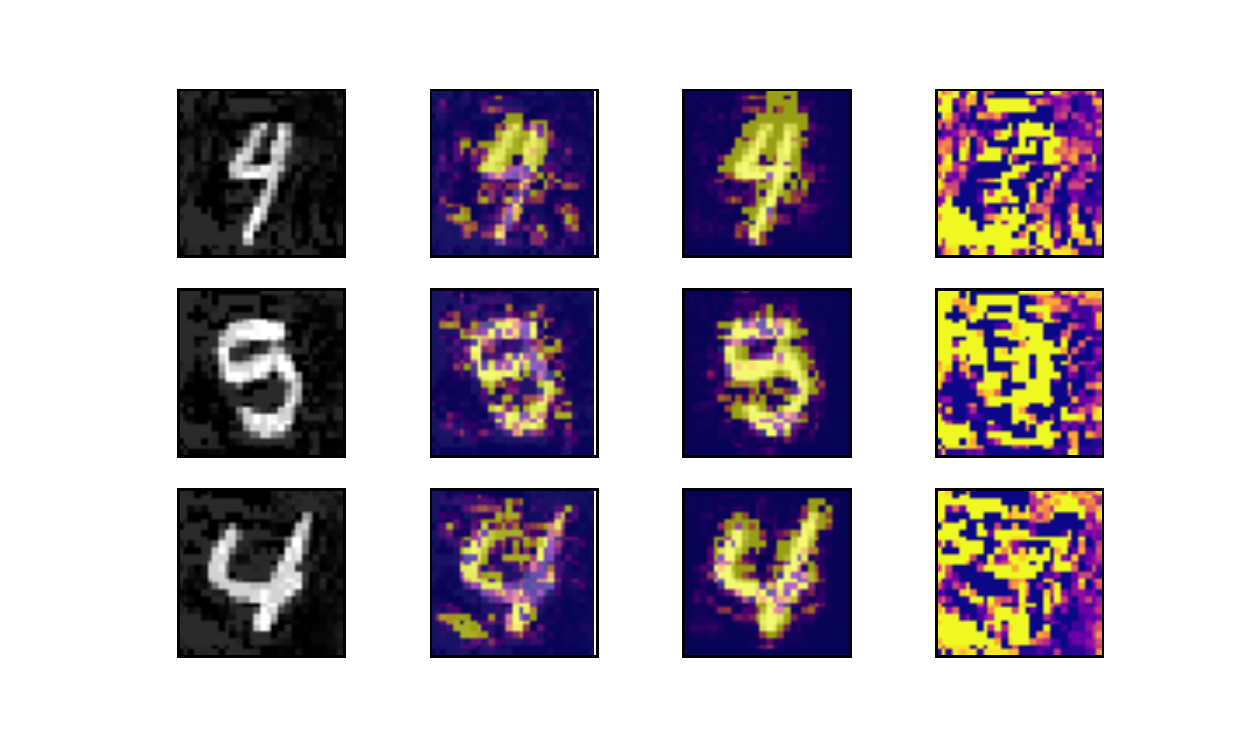}
\subcaption{FIS: Different color scales.}
\label{fig:MNIST_Diff_var}
\end{subfigure}
\caption{Visualization of the Fisher information sensitivity - each row corresponds to one input. The leftmost column shows the adversarial modified image, the second column from the left shows the Fisher information sensitivity for the adversarial image, the third shows the sensitivity for the non-adversarial image using the \emph{same} colorscale on the left image and \emph{different} colorscales on the right image. For comparison the corresponding image was drawn in the background and the difference between adversarial and original image is plotted in the rightmost column. The network classifies the adversarial images from top to bottom on the left as: 8(false), 4(correct) and 3(false) and on the right as: 9(false), 9(correct), 4(correct)}
\end{figure}

We trained a standard convolutional net on the MNIST dataset \cite{LeCun2010}  up to a test accuracy of 98.5\% and a training accuracy of about 99\%. Applying the MI-FGSM \cite{Dong2018} to each image of the test set from with $\epsilon=10^{-1}$ the accuracy drops to 52.5\%. In Figure \ref{fig:MNIST_Hist_ROC} the ROC was presented for the different evaluation techniques \eqref{eq:FisherTrace}, \eqref{eq:FisherForm} and \eqref{eq:NormalizedFisherForm}. For the Fisher form \eqref{eq:FisherForm} and the normalized version \eqref{eq:NormalizedFisherForm} we used the finite difference approximation \ref{eq:FiniteDifference} with $\eps'=10^{-4}$. The curves were computed using 400 randomly selected test images.
The AUC (the area under the ROC curve) is around $92\%$ for all presented techniques and thus pretty decent. Feinman et al. \cite{Feinman2017} for instance, using Bayesian methods and density estimation, report an AUC of 90.6\% for the FGSM and around 97.2\% for the Basic iterative method from \cite{Kurakin2016}. Note however that MNIST is occasionally considered as being too simple for serious adversarial attacks \cite{Carlini2017} and that these attacks usually become efficient for higher dimensions \cite{Goodfellow2014}. In fact, using a fully blown network for detection and neglecting failed adversarial attacks, which we here take into account, Xu et al. \cite{Xu2017} report an AUC of around 99\%. 

In the left Figure \ref{fig:MNIST_Diff} the visualization from the quantity \eqref{eq:FIS} (with normalized $v$) is presented. The leftmost column in each row displays the adversarial image that is studied. The second row from the left  plots the absolute value of \eqref{eq:FIS} for each input node, that is each pixel. The corresponding visualization was made transparent and the adversarial image was plotted in the background. The third column from the right displays the quantities \eqref{eq:FIS} for the input picture without adversarial attack \emph{using the exact same color scale} as for the adversarial picture. The corresponding (non-adversarial) picture was again plotted in the background.  The rightmost column shows the difference between adversarial and the original image as comparison. 

Note that the magnitude of the Fisher information sensitivity \eqref{eq:FIS} is higher for the adversarial examples so that the sensitivity for the non-adversarial images can in fact hardly be noticed. This effect is also true for more complicated problems as MNIST, however we found already for CIFAR10, cf. Section \ref{subsec:cifar10} below, that this striking distinguishability becomes weaker as the problem becomes more involved. 

Let us emphasize that while the Fisher information sensitivity is lower for the non-adversarial image, it is by no means neglectable or meaningless. In the Figure \ref{fig:MNIST_Diff_var} the FIS is plotted for the adversarial and non-adversarial image using \emph{different color scales}, which will also be the convention we will use for CIFAR10 and Fruits-360 below. The area marked by the FIS in Figure \ref{fig:MNIST_Diff_var} is focussed on the digit for the actual input image, while it is only partially aware of the digit for the adversarial image (on the right)  - we will  have a similar finding for more complex problems below. Note finally, that while there seems to be some relation between the area marked for the adversarial image and the pixel difference between adversarial and original image (rightmost image), there is no direct  congruence. If we use individual scaling as in \ref{fig:MNIST_Diff_var}, which we will do from now on, the normalization which lead us to the distinction between \ref{eq:FisherForm} and \eqref{eq:NormalizedFisherForm} looses its importance as it only contributes a scaling factor.

\subsection{CIFAR10}
\label{subsec:cifar10}

\begin{figure}
\begin{subfigure}[t]{0.5\textwidth}
\input{CIFAR10_ROC}
\end{subfigure}
\begin{subfigure}[t]{0.5\textwidth}
\centering
\includegraphics[scale=0.3]{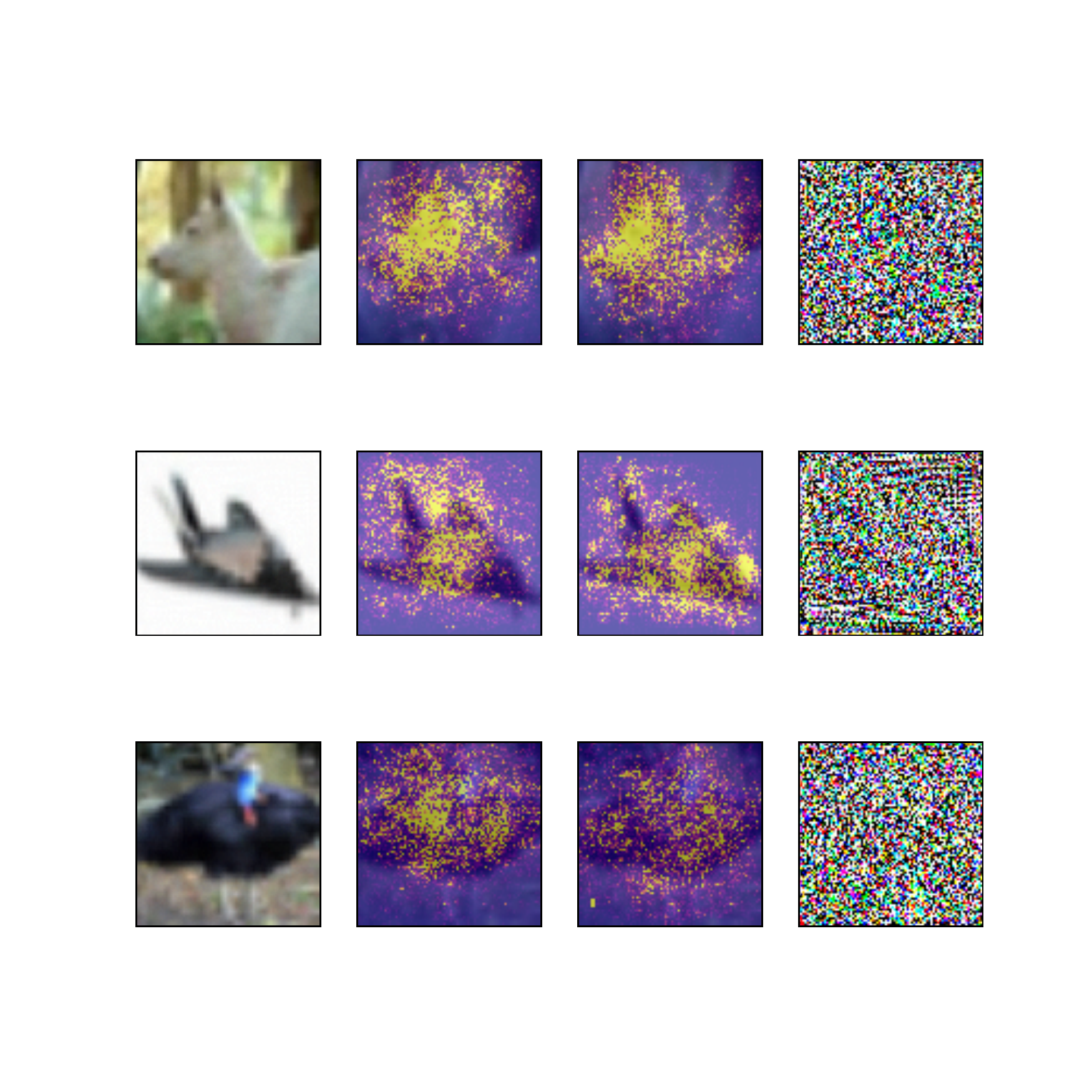}
\end{subfigure}
\caption{\textit{Left}:
ROC for adversarial detection for the CIFAR10 dataset. The AUC is given by 0.79 for the trace of Fisher information  - \eqref{eq:FisherTrace}, 0.77 for the quadratic form in \eqref{eq:FisherForm} and 0.77 for the normalized version \eqref{eq:NormalizedFisherForm}. For details on the adversarial attack compare text. 
\textit{Right}:
FIS visualization for various images from CIFAR10. The structure of this table is the same as in left side of Figure \ref{fig:MNIST_Diff_var}. The labels for deer, airplane and bird were correctly classified by the network on the original data. For the adversarially modified images the network predicted cat, bird and automobile.
}
\label{fig:CIFAR10_ROC_Vis}
\end{figure}

Due to its simplicity and lucidity MNIST is rather easy to interpret when it comes to tasks such as visualization. However, the detection of adversarial examples on more complex problems can be much more challenging and thus more interesting \cite{Carlini2017}. As a next, slightly more evolved, example we consider the CIFAR10 dataset \cite{Krizhevsky2009}. We used a ResNet18 architecture that was pretrained on ImageNet as available in \texttt{torchvision.models} package of pytorch \cite{Paszke2017} and finetuned it to reach a (test-)accuracy of 92.7\% on CIFAR10. To this end images were rescaled to 224$\times$224 prior to feeding through the network.  Due to this higher dimensionality of the input data an adversarial attack tends to be much more effective \cite{Goodfellow2014}. In fact using the MI-FGSM with  $\epsilon=7 \cdot 10^{-3}$ leads the accuracy drop to only $0.2\%$! The left side of Figure \ref{fig:CIFAR10_ROC_Vis} shows the receiver operating characteristic for the quantities introduced in section \ref{sec:detection}. For the computation of the curves we drew 600 randomly selected test images. The AUC for the presented three methods is about $77\%-79\%$ and therefore less as for the much simpler MNIST problem but still markedly above $0.5$ (which would be pure guessing). In \cite{Feinman2017} the authors report an AUC of 72.2\% for the FGSM and of 81\% for the BIM from \cite{Kurakin2016} which is in the same ballpark. 

The visualization using the Fisher information sensitivity from \eqref{eq:FIS} can be seen on the right side of Figure \ref{fig:CIFAR10_ROC_Vis}. For the computation we applied the two-class approximation from \eqref{eq:2class}. The ordering is the same as in Figure \ref{fig:MNIST_Diff_var}, i.e. with \emph{different} color scaling for both sides adapted to the corresponding order of magnitude. We observe once more that the Fisher information sensitivity marks only part of the relevant area for adversarial images (second column from the left), while it covers most of the relevant pixels for the original image (third column from the right). In the middle line for instance the FIS mostly marks the end of the airplane which apparently resembles a bird to the network as this is the prediction on the adversarial image.

\subsection{Fruits-360}

\begin{figure}
\begin{subfigure}[t]{0.5\textwidth}
\input{Fruits_ROC}
\end{subfigure}
\begin{subfigure}[t]{0.5\textwidth}
\includegraphics[scale=0.5]{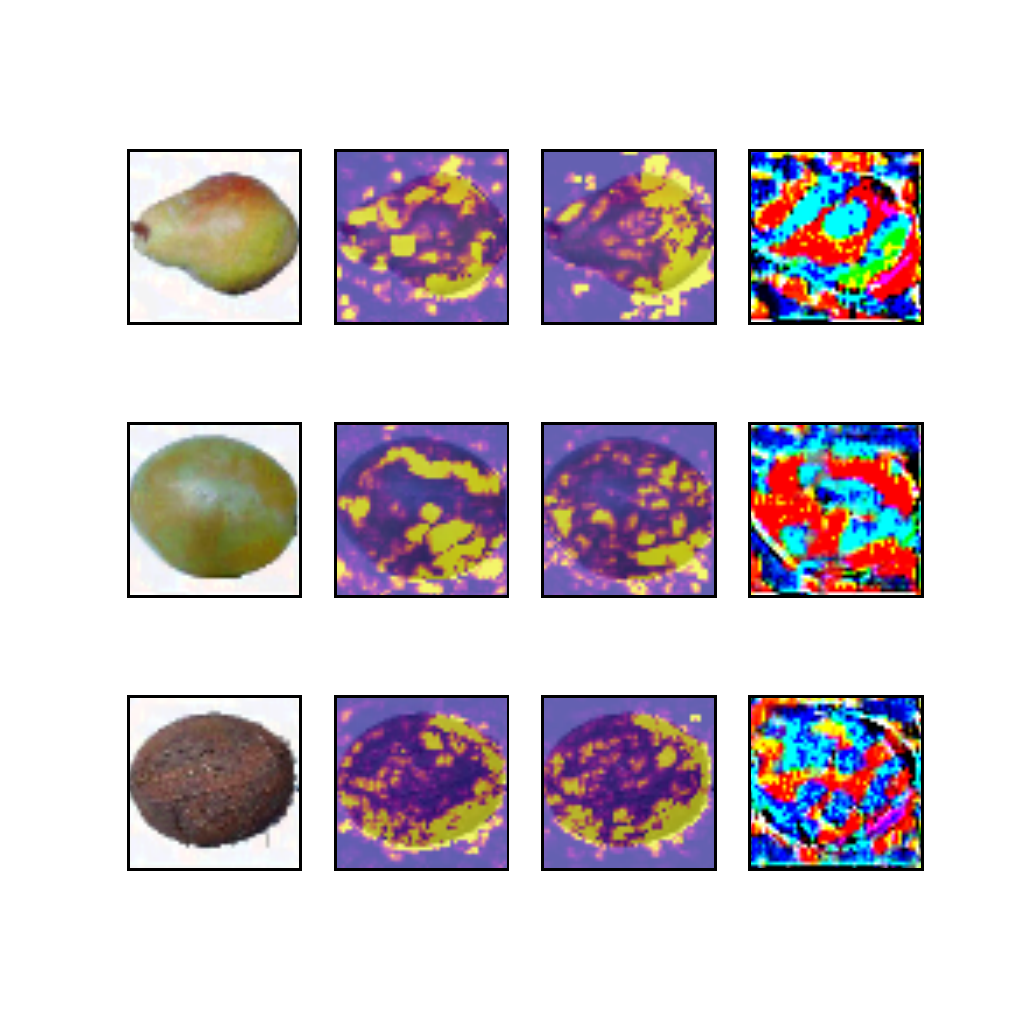}
\end{subfigure}
\caption{\textit{Left}: 
ROC for adversarial detection for the Fruits-360 dataset from \cite{Murecsan2018} that contains 116 different classes of fruits (and vegetables). The AUC is given by 0.86 for the trace of Fisher information  - \eqref{eq:FisherTrace}, 0.82 for the quadratic form in \eqref{eq:FisherForm} and 0.84 for the normalized version from \eqref{eq:NormalizedFisherForm}. For details on the adversarial attack compare text. 
\textit{Right}:
FIS visualization for various images from from the Fruit-360 dataset. The structure of this table is the same as in left side of Figure \ref{fig:MNIST_Diff_var}. The labels for Williams pear, white grape and coconut were correctly classified by the network on the original data. For the adversarial modifications the network predicted pepino melon, banana and chestnut.}
\label{fig:Fruits_ROC_Vis}
\end{figure}

As a final example with a few more classes we consider the Fruit-360 dataset \cite{Murecsan2018} which contains, at the time of the download, 58428 images of resolution $100\times 100$ with fruits (and vegetables) belonging to 116 classes. For validation a test set containing 19932 images is available. Using the same convolutional neural network as proposed in \cite{Murecsan2018} we achieve an accuracy of 98.7\% on the test set. 
We perform an adversarial attack with the MI-FGSM with $\epsilon=7 \cdot 10^{-3}$ that lets the accuracy drop to 12.8\%. 

The left side of Figure \ref{fig:Fruits_ROC_Vis} displays the receiver operating characteristic obtained for the quantities introduced in Section \ref{sec:detection} for a random batch of $600$ images.  For the quadratic form \eqref{eq:FisherForm} and its normalized counterpart \eqref{eq:NormalizedFisherForm} we applied the finite difference approximation from \eqref{eq:FiniteDifference}. The AUC is roughly $82\%$ for the quadratic form \eqref{eq:FisherForm} and around $0.86$ for the normalized Fisher form \eqref{eq:NormalizedFisherForm} and the trace \eqref{eq:FisherTrace}, lying somehow between our result for the MNIST dataset and the one for CIFAR10. 

The visualization based on the FIS is shown in the right-hand side of Figure \ref{fig:Fruits_ROC_Vis}, with the same scheme as in Figure \ref{fig:MNIST_Diff_var}, that is with color scaling adapted to the corresponding order of magnitude. The FIS was computed using the two-class approximation from \eqref{eq:2class}. While for some images, compare the last row for instance, the FIS marks pretty similar areas for the adversarial and the original image there are interesting differences for others. Consider the second row, which is for a picture that actually shows a white Grape. While the network makes the correct classification for the original image, it classifies the adversarial image as being a banana, which might correspond to the curvy structures that are highlighted by the FIS. 

\section{Conclusions}
We introduced several quantities that are based on the Fisher information and can be used to detect adversarial examples. In contrast to the Fisher information these quantities are tractable even for larger network sizes. The effectiveness of these quantities was studied on various datasets of different complexity. A usage of the proposed concept based on the Fisher information for the study of adversarial attacks seems natural as it measures the ``curvature'' on a statistical manifold and thus the compliance between input and learned parameters. 

From two of these quantities, namely \eqref{eq:FisherForm} and \eqref{eq:NormalizedFisherForm} we can derive objects that can judge the influence of each input node on the Fisher information, this can be used for visualization purposes. Applying this tool to adversarial and non-adversarial images leads to the highlighting of different areas. 

The presented techniques are not restricted to the usage with adversarial examples but can also be used in a general context to raise a red flag on an unusual input. Using the presented visualization technique the decision of the network can further be investigated and therefore be used for a sanity check. 

\bibliographystyle{acm}
{
\scriptsize
\bibliography{Fisher_Adversarial}}

\end{document}